\documentclass[10pt, conference]{IEEEtran}
\IEEEoverridecommandlockouts

\usepackage{cite}
\usepackage{amsmath,amssymb,amsfonts}
\usepackage{algorithmic}
\usepackage{graphicx}
\usepackage{textcomp}
\usepackage{xcolor}
\usepackage{url}
\usepackage[font=normalsize]{caption}

\def\BibTeX{{\rm B\kern-.05em{\sc i\kern-.025em b}\kern-.08em
    T\kern-.1667em\lower.7ex\hbox{E}\kern-.125emX}}

\begin{document}

\title{Benchmarking Vector, Graph and Hybrid Retrieval Augmented Generation (RAG) Pipelines for Open Radio Access Networks (ORAN)}

\author{\IEEEauthorblockN{Sarat Ahmad, Zeinab Nezami, Maryam Hafeez, Syed Ali Raza Zaidi}
\IEEEauthorblockA{\textit{School of Electronic and Electrical Engineering}, \textit{University of Leeds, UK}\\
\{S.Ahmad, Z.Nezami, M.Hafeez, S.A.Zaidi\}@leeds.ac.uk}
}

\maketitle

\begin{abstract}
Generative AI (GenAI) is expected to play a pivotal role in enabling autonomous optimization in future wireless networks. Within the ORAN architecture, Large Language Models (LLMs) can be specialized to generate xApps and rApps by leveraging specifications and API definitions from the RAN Intelligent Controller (RIC) platform. However, fine-tuning base LLMs for telecom-specific tasks remains expensive and resource-intensive. Retrieval-Augmented Generation (RAG) offers a practical alternative through in-context learning, enabling domain adaptation without full retraining. While traditional RAG systems rely on vector-based retrieval, emerging variants such as GraphRAG and Hybrid GraphRAG incorporate knowledge graphs or dual retrieval strategies to support multi-hop reasoning and improve factual grounding. Despite their promise, these methods lack systematic, metric-driven evaluations, particularly in high-stakes domains such as ORAN. In this study, we conduct a comparative evaluation of Vector RAG, GraphRAG, and Hybrid GraphRAG using ORAN specifications. We assess performance across varying question complexities using established generation metrics: faithfulness, answer relevance, context relevance, and factual correctness. Results show that both GraphRAG and Hybrid GraphRAG outperform traditional RAG. Hybrid GraphRAG improves factual correctness by 8\%, while GraphRAG improves context relevance by 11\%.
\end{abstract}


\section{Introduction}
RAG has emerged as a transformative advancement for enhancing LLMs in the telecommunications domain. By enabling dynamic retrieval of domain-specific knowledge, RAG facilitates the generation of fact-based, contextually relevant responses~\cite{yilma2025telecomrag}. This is especially valuable in telecommunications, where the complexity and rapid evolution of standards, protocols, and specifications necessitate not only accurate retrieval but also coherent, relevant, and verifiably grounded responses~\cite{bornea2024telco}.

As the RAG architecture continues to evolve, recent advancements have introduced more structured retrieval strategies that go beyond simple lexical and vector-based similarity search to more sophisticated approaches such as multi-level graph based retrieval~\cite{chen2024knowledge}. Within this context, GraphRAG~\cite{edge2024local} has emerged as a promising paradigm that organises information into knowledge graphs (KGs) and leverages graph traversal techniques to retrieve contextually relevant subgraphs in response to queries. This structure enables the model to produce more nuanced, connected, and semantically grounded responses. Moreover, GraphRAG enables the structuring of implicit knowledge by relating entities across multiple datasets, supporting advanced capabilities such as multi-hop reasoning and both global and local summarization~\cite{edge2024local}.

Recent research has proposed Hybrid GraphRAG~\cite{sarmah2024hybridrag}, a unified framework that combines vector-based and graph-based retrieval to leverage the complementary strengths of semantic similarity and structured reasoning. This fusion has been shown to enhance factuality and completeness in domain-specific applications such as finance, healthcare, and cybersecurity~\cite{zhang2025survey}. However, the evaluation of these systems in the telecommunications domain, particularly ORAN, remains unexplored.

Evaluating these systems is particularly important in modern telecom environments. RAG-based implementations support a range of advanced use cases, including xApp/rApp generation via in-context learning with telecom-specific LLMs~\cite{wu2025llm}, root cause analysis using knowledge graphs constructed by GraphRAG~\cite{yuan2025enhancing}, and intent-driven network management through the generation of Infrastructure-as-Code (IaC) or Configuration-as-Code (CaC)~\cite{dzeparoska2023llm}. GraphRAG and Hybrid GraphRAG architectures demonstrate strong potential in these scenarios by enabling multi-hop reasoning across configuration constraints, interface specifications, and data privacy policies~\cite{xiong2024graph}. Therefore, we address the broader challenge of conducting a systematic and open evaluation of these retrieval-augmented architectures.

Traditional evaluation approaches, which rely on coarse metrics such as Precision, Recall, ROUGE, and F1-scores, often fail to capture critical dimensions of response quality, including contextual alignment and factual grounding~\cite{yu2024evaluation}. To overcome these limitations, we adopt LLM-based evaluation methods, which have shown strong potential as reference-free evaluators capable of approximating human judgment with high reliability~\cite{zheng2023judging}. By employing independent generation metrics such as faithfulness, answer relevance, and context relevance, we provide a more comprehensive and interpretable assessment of system performance~\cite{es2024ragas}, particularly in high-
stakes applications such as ORAN.

In this work, we present an open, metric-driven comparison of Hybrid GraphRAG, GraphRAG, and Vector RAG pipelines using ORAN specification documents. Each system is evaluated across both structured and unstructured Question Answering (QA) tasks, with performance analyzed across varying levels of question complexity. Our main contributions are as follows:
\begin{enumerate} \item \textbf{Three-Way Open Evaluation:} We conduct a side-by-side comparison of Vector RAG, GraphRAG, and Hybrid GraphRAG pipelines on ORAN specification documents. To support transparency and reproducibility, the complete pipeline setup and evaluation code is available on github: https://github.com/cheddarhub/rag-eval-oran

\item \textbf{Metric-Driven Comparative Analysis:} We employ independent generation metrics: faithfulness, answer relevance, context relevance, and factual correctness, assessing across critical dimensions of response quality. 

\item \textbf{Complexity-Aware Performance Insights:} Using the ORAN-13K benchmark~\cite{gajjar2024oran}, we analyze how each model performs under varying reasoning demands, providing insight into performance trade-offs across different levels of question complexity.\end{enumerate}

\section{Related Work}
A baseline RAG model~\cite{lewis2020retrieval} consists of two key components: a retriever and a generator. The retriever selects semantically similar context from a vector-based knowledge database, which is then combined with the query and passed to an LLM-based generator to produce a coherent, context-aware response. The evaluation of RAG systems has evolved from traditional metrics such as ROUGE and BLEU, focused on lexical overlap, to LLM-based evaluators that enable context-aware assessments of coherence, fluency, and relevance~\cite{yu2024evaluation}. A key contribution is the RAGAS framework~\cite{es2024ragas}, which offers automated, reference-free evaluation using independent LLM-based metrics such as faithfulness, answer relevance, and context relevance. Roychowdhury et al.~\cite{roychowdhury2024evaluation} enhance this approach by introducing greater transparency through intermediate output capture and prompt engineering for domain-specific tasks, including telecom adaptation. Other works, including ARES~\cite{saad2023ares} and RAGEval~\cite{zhu2024rageval}, further advance automated RAG evaluation using contrastive learning, lightweight LLMs, and automatic dataset generation.

\begin{figure}[!htbp]
\centering
\includegraphics[width=8.5cm, 
height=7cm,
trim=11.3cm 6.5cm 13.2cm 4.9cm,clip
]{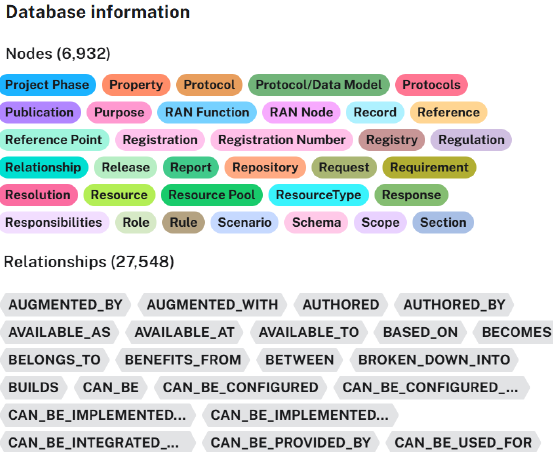}
\caption{A section of the graph database, showing the distribution of node categories and relationship types.}
\label{database}
\end{figure}

GraphRAG~\cite{edge2024local} extends RAG by structuring information into hierarchical KGs, enabling multihop reasoning and revealing implicit relationships across data. Its transparent path traceability makes it particularly effective for domain-specific tasks requiring complex reasoning and specialized terminology. However, while standard RAG pipelines have been widely evaluated using transparent metrics such as RAGAS, systematic evaluation of GraphRAG remains limited, especially in the telecommunications and ORAN context.

Several recent studies~\cite{han2025rag,wang2025causalrag} have explored GraphRAG evaluation, though primarily with conventional metrics. Han et al.~\cite{han2025rag} present one of the first direct comparisons between GraphRAG and RAG on general benchmark datasets but restrict their analysis to coarse metrics such as Precision, Recall, and F1-score. Wang et al.~\cite{wang2025causalrag} apply RAGAS-style evaluation but focus on generalized datasets, without addressing domain-specific terminology or relational complexity.

The Hybrid GraphRAG approach combines vector-based semantic retrieval with graph traversal to balance broad document coverage with structured, relationship-rich context~\cite{sarmah2024hybridrag}. Sarmah et al. evaluate RAG, GraphRAG, and Hybrid GraphRAG within the finance domain, demonstrating performance gains through this integration. In wireless networking, Xiong et al.~\cite{xiong2024graph} assess RAG and GraphRAG using open metrics on raw network data, though their work does not target ORAN specifications. Similarly, SMART-SLIC~\cite{barron2024domain} applies hybrid retrieval in cybersecurity using non-negative tensor factorization and automated graph construction without LLM dependence. While these studies highlight the potential of hybrid retrieval across domains, systematic and transparent evaluation of Vector RAG, GraphRAG, and Hybrid GraphRAG within the ORAN context remains unexplored.\\ 
Our work addresses this gap by presenting a direct, metric-driven comparison of these approaches using ORAN specification data.

\begin{figure*}[!htbp]
\centering
\includegraphics[width=\textwidth, 
trim=4.8cm 7.3cm 2.5cm 2cm,clip
]{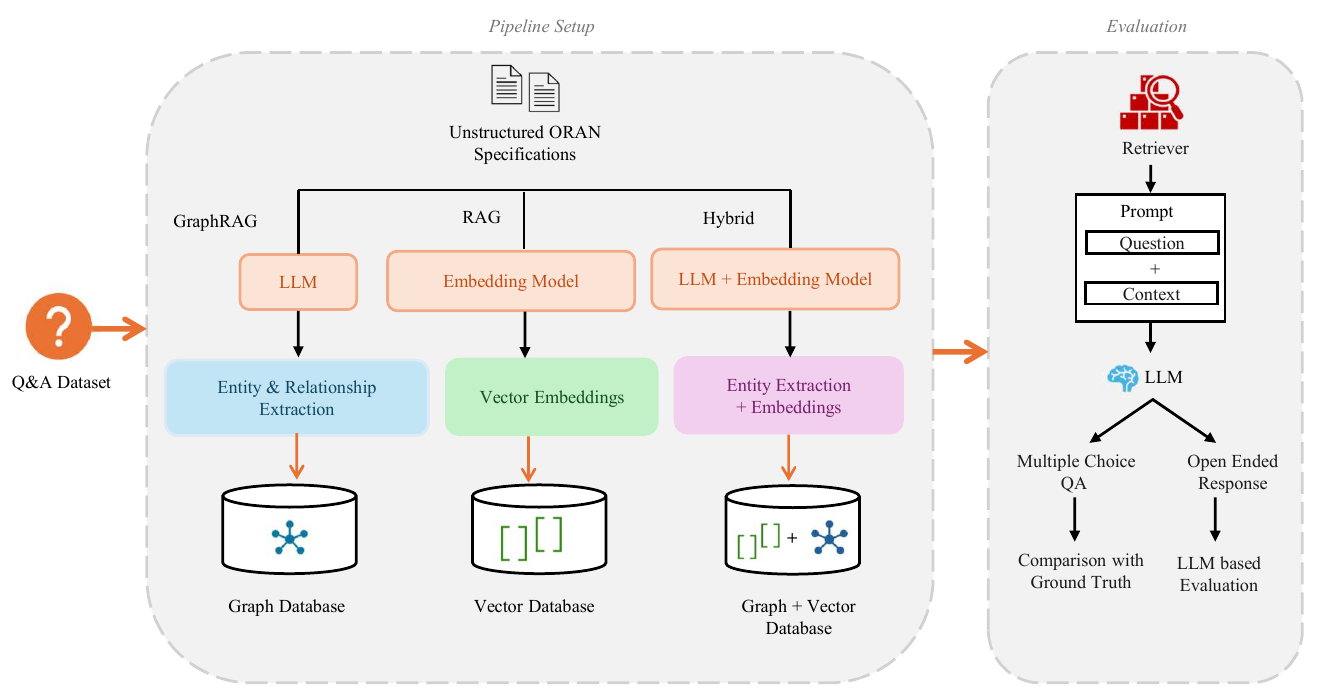}
\caption{Overview of the experimental pipeline, illustrating the core components and data flow across the retrieval, generation and evaluation stages.}
\label{overview}
\vspace{-0.1 in}
\end{figure*}

\section{System Design}
Figure~\ref{overview} illustrates the overall architecture of our experimental setup, which includes three pipelines: GraphRAG, Vector RAG and Hybrid GraphRAG. This section outlines the design and operational workflow of each pipeline, the dataset used, and the experimental configuration adopted to ensure a fair and consistent comparison.

\subsection{GraphRAG}
We employ the \textit{Neo4j LLM Knowledge Graph Builder}~\cite{Hoppa_2025} to construct a knowledge graph from unstructured textual data.  The input documents are initially processed by the \textit{LLMGraphTransformer}~\cite{LLMGraphTransformer}, which extracts entities and their semantic relationships, thereby transforming raw text chunks into a structured graph representation. In the constructed knowledge graph, nodes represent entities, while edges denote the relationships among them. The graph is then stored in \textit{Neo4j AuraDB}~\cite{neo4j_aura_db}, enabling efficient storage and retrieval via Cypher queries. Figure~\ref{database} shows a snapshot from the graph database. To retrieve relevant context, an entity extraction chain is employed to identify key entities from the input query. Table~\ref{entity_examples} provides details on entities extracted from the query based on a pre-defined schema. The identified entities are then used to construct a Cypher query, which traverses the knowledge graph to retrieve associated nodes and their relationships.

\begin{table}[htbp]
\caption{Entity Categories for Structured Query Extraction}
\vspace{-0.1 in}
\label{entity_examples}
\small
\begin{center}
\renewcommand{\arraystretch}{1.3}
\begin{tabular}{p{2cm}p{5cm}}
\hline
\textbf{Category} & \textbf{Description} \\
\hline
    Organisations & Organisations/Alliances mentioned (e.g., ORAN Alliance, 3GPP). \\
    Architecture & Network functions and architectural elements (e.g., SMO, Near-RT RIC, NSSMF, DU, CU, RU). \\
    Standards & Protocols, standards, and interfaces (e.g., E2AP, O1, A1, TS 38.401, WG1, WG6). \\
    Technology & Technologies and use cases (e.g., AI/ML, cloud-native, energy efficiency, SLA assurance). \\
    References & Document identifiers and other telecom-specific terms or APIs (e.g., ORAN.WG1.SPEC-2023-v06.00). \\
\hline
\end{tabular}
\label{tab1}
\end{center}
\end{table}

\subsection{Vector RAG}
For the RAG pipeline, unstructured textual data is initially loaded using the \textit{PyPDFLoader}~\cite{langchain_pypdfloader} utility and then segmented into smaller units using LangChain’s \textit{Recursive Text Splitters}~\cite{recursive_character_text_splitter}. These text chunks are subsequently embedded and the resulting vector representations are stored in a Chroma vector database~\cite{chroma2025} to facilitate similarity-based retrieval.  During inference, the input query is embedded and compared to stored vectors using cosine similarity. The top four matching chunks are retrieved and combined with the query and a tailored prompt, which is then sent to the LLM for response generation.

\subsection{Hybrid GraphRAG}
The Hybrid GraphRAG technique integrates both vector-based retrieval and graph-based traversal to leverage the strengths of each approach. Initially, semantic similarity search is used to retrieve relevant text chunks, followed by graph traversal to extract structured, relationship-rich information. The retrieved contexts are then concatenated, placing the vector-based content first, followed by the graph-derived context. The final prompt is designed to guide the language model to prioritize information from Vector RAG for generating a broad and comprehensive answer, while using the GraphRAG context to supplement with structural details and domain-specific relationships.

\begin{figure*}[htbp]
\centering
\includegraphics[
width=\textwidth, 
trim=11cm 11.6cm 2cm 2.6cm,clip
]{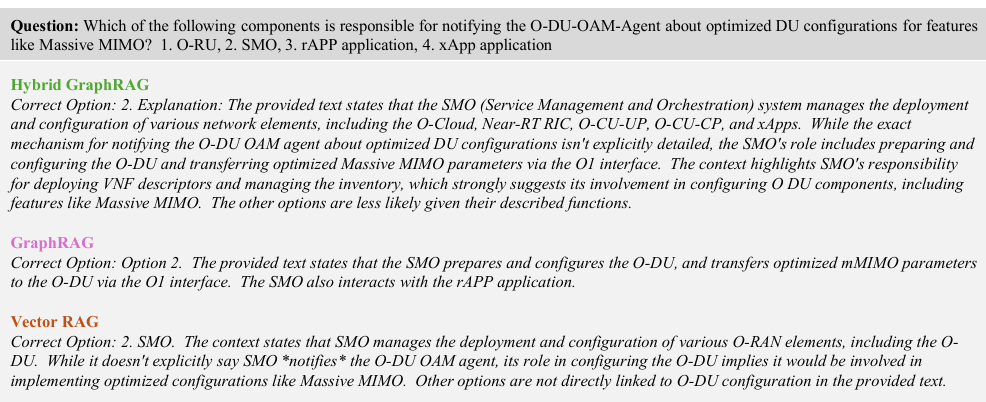}
\caption{Example response generated by Vector RAG, GraphRAG, and Hybrid GraphRAG for a benchmark question from the ORAN-Bench-13K dataset.}
\label{answers}
\end{figure*}

\subsection{Dataset}
For this study, we utilized a corpus of 74 documents from the ORAN Alliance Specifications~\cite{Developed_by_HAVIT}. The evaluation was conducted using the \textit{ORAN-Bench-13K} dataset~\cite{gajjar2024oran}, a benchmark specifically designed to assess the performance of LLMs within the ORAN context. The dataset categorizes questions into three levels of complexity: Easy, Intermediate, and Hard. We selected a stratified subset of 600 questions to ensure balanced coverage across all difficulty levels and representative topical diversity. The sampling procedure maintained proportional representation from categories such as network architecture, analytics and monitoring, anomaly detection, and protocol interpretation. Each question includes four answer options along with a ground truth label. The categorization reflects increasing levels of difficulty: (i) Easy: Questions targeting foundational concepts or factual knowledge (Simple QA) (ii) Intermediate: Questions requiring moderate reasoning, comprehension, or application of concepts (Complex Reasoning QA) (iii) Hard: Questions demanding deep understanding of ORAN standards and the ability to synthesize information across multiple documents (Multi-hop Reasoning) ~\cite{zhang2025survey}.

\subsection{Comparison Setup and Configuration}
To evaluate the pipelines, the benchmark dataset was processed to generate structured outputs comprising the question, answer options, retrieved context, generated response, predicted answer, and ground truth. To ensure fair comparison across all three pipelines, experimental parameters were held constant. The generator model used was \textit{Gemini 1.5 Flash}~\cite{team2024gemini}, with \textit{models/embedding-001} as the embedding model for consistent semantic representation. Context chunks were generated with a size of 1024 tokens and no overlap, retrieving the top four most relevant chunks per query based on cosine similarity. This setup isolates the retrieval strategy as the primary variable, ensuring the validity and comparability of the evaluation results across all methods.

\section{Evaluation}

\begin{figure*}[tp]
\centering
\includegraphics[width=\textwidth,
]{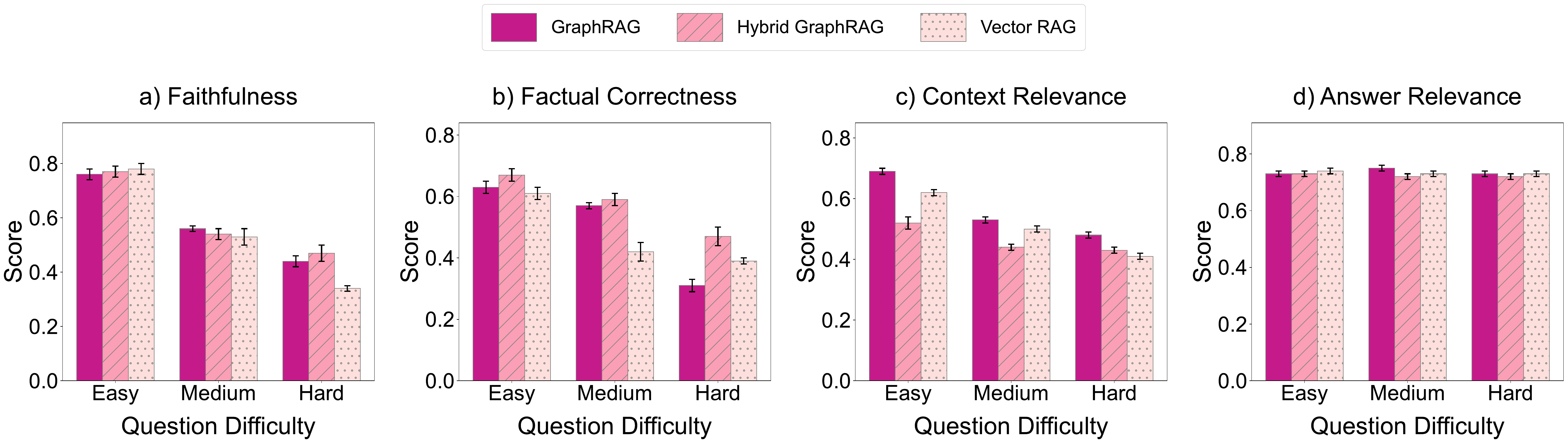}
\caption{Comparison of Vector RAG, GraphRAG, and Hybrid GraphRAG Across Question Difficulty Levels for Four Evaluation Metrics.}
\label{Graphs}
\vspace{-0.1 in}
\end{figure*}

To comprehensively assess the effectiveness of each retrieval pipeline, we adopt a dual evaluation strategy. The first approach involves multiple-choice question answering using a benchmark dataset with ground truth answers. The second approach assesses open-ended generation, where LLMs act as evaluators to compute metric scores. This LLM based evaluation allows end-to-end assessment across both retrieval and generation stages without explicit separation, facilitating automated scoring of critical response attributes~\cite{zou2024telecomgpt, sarmah2024hybridrag}. 

\subsection{Evaluation Metrics}
We adopt a set of reference-free, LLM-based metrics from the RAGAS framework~\cite{es2024ragas} to capture key aspects of response quality. These include:

\subsubsection{\noindent{Faithfulness}}Measures the extent to which the generated response is grounded in the retrieved context. The evaluation involves (i) Statement Decomposition using an LLM to extract verifiable statements from the response and (ii) Statement Verification against the retrieved context. The final faithfulness score, \( F \), is then computed as \( F = \frac{|V|}{|S|} \), where \( |V| \) is the number of statements that were verifiable and \( |S| \) is the total number of statements.
\subsubsection{\noindent{Answer Relevance}}Assesses how well the generated response addresses the original question, independent of factual correctness. Multiple questions are generated from the open-ended response using an LLM. Embeddings for the original and generated questions are computed, and semantic similarity is measured via cosine similarity. The final score is the average similarity across these pairs,
given by \( AR = \frac{1}{n} \sum_{i=1}^{n} \text{sim}(q, q_i) \), where \( \text{sim}(q, q_i) \) denotes the cosine similarity between the embedding of the original question \( q \) and each of the \( n \) generated questions \( q_i \).
\subsubsection{\noindent{Context Relevance}}Measures how well the retrieved context focuses on the information necessary to answer the given question, penalizing redundant or irrelevant content. An LLM extracts key sentences from the context that directly support answering the question. The score is calculated as the ratio of relevant extracted sentences to the total number of sentences in the context.
\subsubsection{\noindent{Factual Correctness}}Measures the accuracy of the predicted answer relative to the ground truth in the MCQ setting. The score is the ratio of correctly predicted answers to the total number of questions in the dataset.

\subsection{Results}
Figure~\ref{Graphs} visualizes the performance of each model across four core evaluation metrics stratified by question difficulty. To provide a broader perspective, Table~\ref{Average} reports the mean of each evaluation metric aggregated across easy, medium, and hard questions. The corresponding standard deviations capture variability in performance across these difficulty levels. Each experiment is repeated thrice to ensure consistency and reliability. The results are discussed below, organized by evaluation metric.

\begin{table}[htbp]
\caption{Average Performance Across Question Difficulty Levels}
\label{Average}
\small
\begin{center}
\renewcommand{\arraystretch}{1.5}
\begin{tabular}{@{}lccc@{}}
\hline
\textbf{Metric} & \textbf{Graph} & \textbf{Hybrid} & \textbf{Vector} \\
\hline
    Faithfulness & 0.59 $\pm$ 0.16 & 0.59 $\pm$ 0.16 & 0.55 $\pm$ 0.22 \\
    Factual Correctness & 0.50 $\pm$ 0.17 & 0.58 $\pm$ 0.10 & 0.48 $\pm$ 0.12 \\
    Context Relevance & 0.56 $\pm$ 0.10 & 0.45 $\pm$ 0.05 & 0.51 $\pm$ 0.11 \\
    Answer Relevance & 0.74 $\pm$ 0.01 & 0.72 $\pm$ 0.01 & 0.73 $\pm$ 0.01 \\
\hline
\end{tabular}
\label{tab1}
\end{center}
\end{table}

In terms of faithfulness, both GraphRAG and Hybrid GraphRAG outperform Vector RAG by 4\%, with lower variability across difficulty levels. This suggests that the graph-based pipelines produce responses that are more consistently grounded in the retrieved context and are less susceptible to hallucinations.

In terms of factual correctness, Hybrid GraphRAG achieves the highest average score with low variability (0.58 ± 0.10) across all difficulty levels, as shown in Table~\ref{Average}. Its performance remains stable due to its ability to compensate for limitations in individual retrieval strategies, falling back on vector-based retrieval when graph-derived context is insufficient, and vice versa. GraphRAG follows with an average score of 0.50, limited by its dependence on entities and relationships extracted from the knowledge graph, which may be incomplete or sparse. As shown in Figure~\ref{Graphs}, Vector RAG performs best on easy questions (0.61), but its accuracy drops on medium and hard questions due to its reliance on direct semantic similarity, which is less effective when relevant information is not explicitly retrieved.

With respect to context relevance, GraphRAG outperforms Hybrid GraphRAG by 11\% on average across all question difficulty levels. This improvement is attributed to GraphRAG’s use of structured entities and relationships, which facilitates the retrieval of concise information while minimizing irrelevant content. In contrast, Hybrid GraphRAG exhibits the lowest context relevance score (0.45 ± 0.05), indicating consistently weaker alignment between retrieved context and query across difficulty levels. As shown in Figure~\ref{answers}, the response generated by Hybrid GraphRAG, though factually accurate, tends to include verbose or tangential details that dilute relevance and reduce semantic precision with respect to the query. Answer relevance remains consistent across models, with GraphRAG showing a slight lead (0.74 ± 0.01), likely due to its structured graph traversal that promotes more focused responses.

These findings offer insights for deploying RAG systems in telecom environments. Hybrid GraphRAG is well-suited for reasoning-intensive tasks such as xApp/rApp generation or federated orchestration, where completeness is prioritized. GraphRAG, with its focused and concise outputs, is better aligned with latency-sensitive applications like root cause analysis or intent-driven network management. Overall, the trade-offs in retrieval precision, verbosity, and efficiency underscore the need to align RAG architectures with specific performance and operational requirements in ORAN use cases.

\section{Conclusion}
This study presents a systematic evaluation of Vector RAG, GraphRAG, and Hybrid GraphRAG within the telecommunications domain using ORAN specification data. 
Each pipeline is assessed across varying question complexities using LLM based, independent generation metrics. Results indicate that GraphRAG and Hybrid GraphRAG outperform Vector RAG on complex reasoning tasks. GraphRAG achieves superior context and answer relevance while Hybrid GraphRAG demonstrates higher factual correctness, albeit with increased redundancy and computational cost. Future work could explore  empirically quantifying latency and compute overhead across pipelines and integrating multimodal context into the retrieval pipeline to enhance reasoning for dynamic telecom tasks. Additionally, deploying these pipelines within orchestration frameworks (e.g., SMO or RIC) would enable practical evaluation for use cases such as intent-based service provisioning.

\section{Acknowledgment}
This research was supported by UK Research and Innovation (UKRI) through the EPSRC under two grants: the Technology Missions Fund project CHEDDAR (EP/Y037421/1), and Award UKRI851, focused on strategic decision-making and cooperation among AI agents in telecom safety and governance. This study does not involve human subjects or sensitive data, and raises no ethical or policy concerns

\bibliographystyle{IEEEtran}
\bibliography{reference}
\end{document}